\pdfoutput=1

\documentclass[11pt]{article}

\usepackage{ACL2023}
\usepackage{tablefootnote}

\usepackage{times}
\usepackage{latexsym}
\usepackage{graphicx} 
\usepackage[T1]{fontenc}

\usepackage[utf8]{inputenc}

\usepackage{microtype}

\usepackage{inconsolata}
\usepackage{multirow}
\usepackage{booktabs}
\usepackage{url}
\usepackage{amsfonts}
\usepackage{amsmath}

\usepackage{algpseudocode}
\usepackage[vlined, ruled,linesnumbered]{algorithm2e}

\newcommand{\Es}{\color{red}}

\renewcommand{\Es}{}

\title{M-RAG: Reinforcing Large Language Model Performance through Retrieval-Augmented Generation with Multiple Partitions}

\author{Zheng Wang\textsuperscript{1}, Shu Xian Teo\textsuperscript{1}, Jieer Ouyang\textsuperscript{1}, Yongjun Xu\textsuperscript{1}, Wei Shi\textsuperscript{1}\\
  \textsuperscript{1}Huawei Technologies, Co., Ltd.\\
\texttt{\{wangzheng155,teo.shu.xian,ouyang.jieer,xuyongjun6,w.shi\}@huawei.com}}

\begin{document}
\maketitle
\begin{abstract}

Retrieval-Augmented Generation (RAG) enhances Large Language Models (LLMs) by retrieving relevant memories from an external database. However, existing RAG methods typically organize all memories in a whole database, potentially limiting focus on crucial memories and introducing noise. In this paper, we introduce a multiple partition paradigm for RAG (called \texttt{M-RAG}), where each database partition serves as a basic unit for RAG execution. Based on this paradigm, we propose a novel framework that leverages LLMs with Multi-Agent Reinforcement Learning to optimize different language generation tasks explicitly. Through comprehensive experiments conducted on seven datasets, spanning three language generation tasks and involving three distinct language model architectures, we confirm that \texttt{M-RAG} consistently outperforms various baseline methods, achieving improvements of 11\%, 8\%, and 12\% for text summarization, machine translation, and dialogue generation, respectively.

\end{abstract}

\section{Introduction}
\label{sec:introduction}

Introduced by~\cite{lewis2020retrieval}, Retrieval-Augmented Generation (RAG) represents a paradigm within the domain of Large Language Models (LLMs) to augment generative tasks. More specifically, RAG incorporates an initial retrieval step where LLMs query an external database to acquire relevant information before progressing to answer questions or generate text. This process not only guides the subsequent generation step but also guarantees that the responses are firmly anchored in the retrieved information (referred to as memories). Consequently, it enhances LLM performance, and has attracted growing research interests~\cite{DBLP:journals/corr/abs-2312-10997} in recent years.

While the majority of existing studies~\cite{DBLP:journals/corr/abs-2310-11511,cheng2023lift,DBLP:journals/corr/abs-2305-14283} adopt a retrieval approach that considers \emph{a database as a whole}, which tends to yield a coarse-grained retrieval. The collective organization of all memories may hinder the focus on crucial memories and introduce noise, particularly due to the inherent challenges of Approximate k-Nearest Neighbor (AKNN) search when applied to large datasets. In this context, we investigate a retrieval approach that aims to search within a partition of the database, corresponding retrieval at a fine-grained level, which is designed to enhance the generation process by targeting specific memories. 
Moreover, in quite a few vector database systems, database partitions are regarded as fundamental units for analysis. This facilitates the construction and maintenance of index structures~\cite{pan2023survey}, ensures the protection of user privacy data (stored in specific partitions with access rights)~\cite{xue2017secure}, and supports distributed architectures~\cite{guo2022manu}. Therefore, in this work, we propose to take \emph{a partition as a basic entity} in the execution of RAG, which is less explored in current methods.

\begin{figure*}[ht]
	\centering
	\hspace*{-4mm}
	\begin{tabular}{c c c c}
 	\hspace{-3mm}
		\begin{minipage}{0.24\textwidth}
			\includegraphics[width=\textwidth]{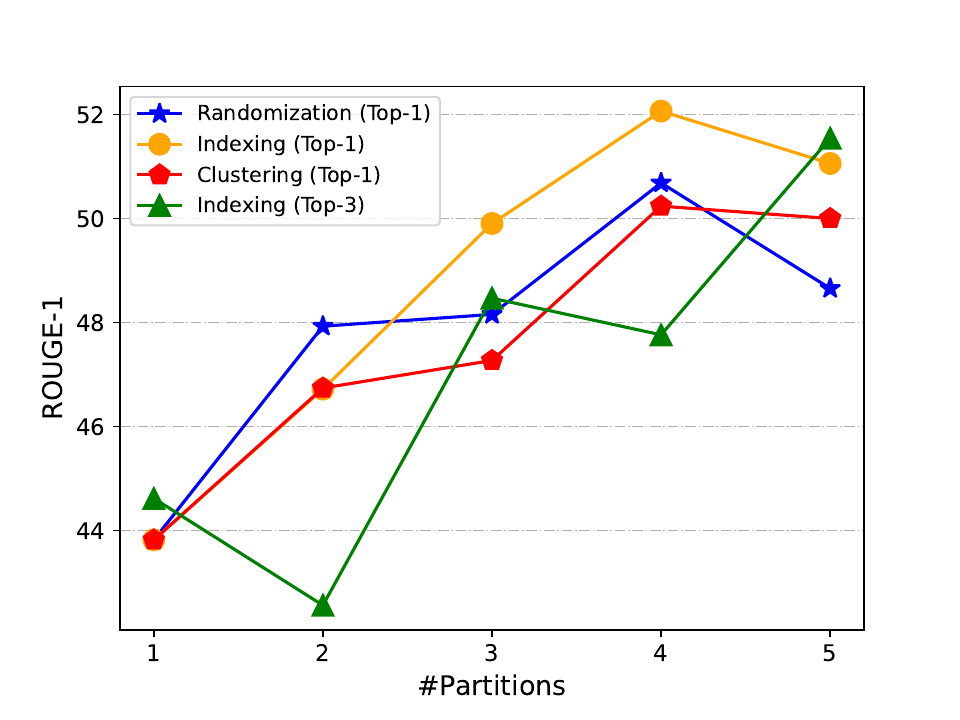}
		\end{minipage}\hspace{-3mm}
		&
		\begin{minipage}{0.24\textwidth}
			\includegraphics[width=\textwidth]{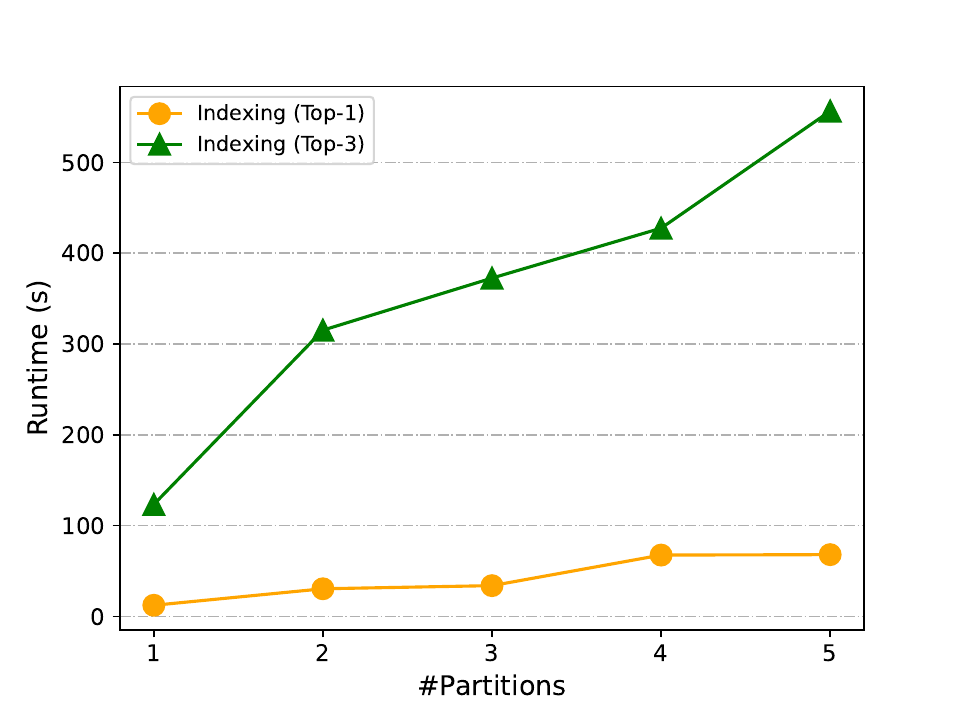}
		\end{minipage}\hspace{-3mm}
		&
		\begin{minipage}{0.24\textwidth}
			\includegraphics[width=\textwidth]{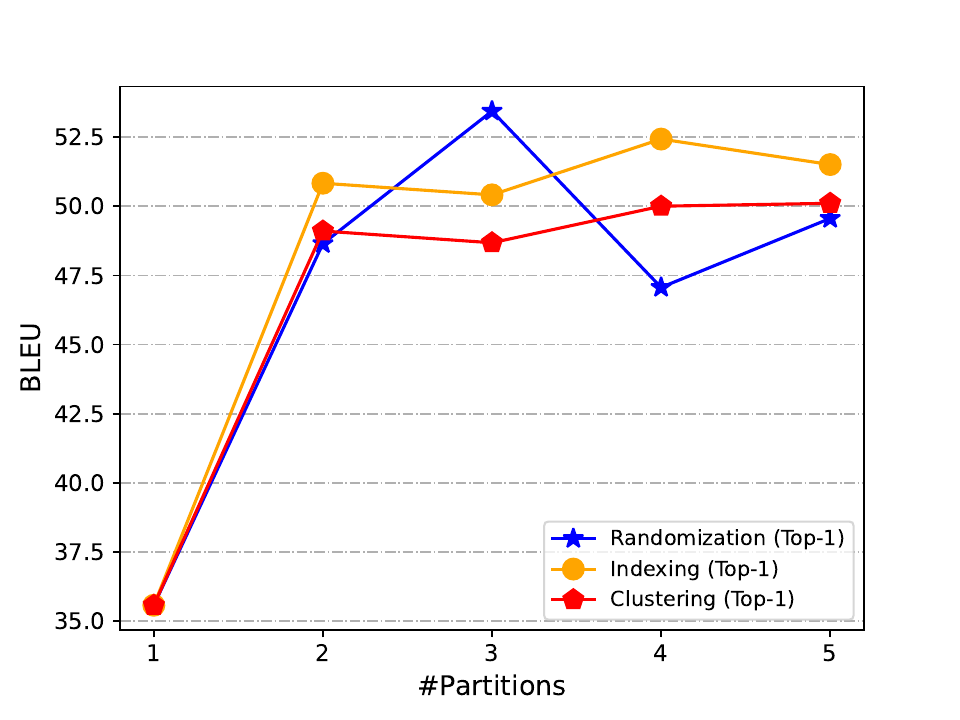}
		\end{minipage}\hspace{-3mm}
		&
		\begin{minipage}{0.24\textwidth}
			\includegraphics[width=\textwidth]{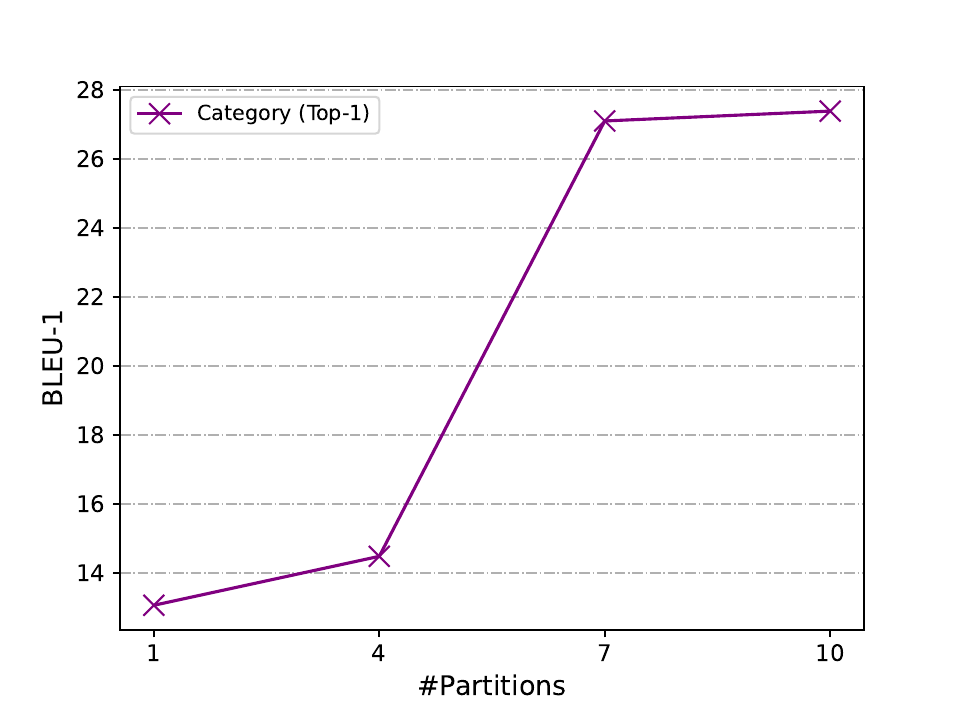}
		\end{minipage}\hspace{-3mm}
		\\
		(a) Summ. (ROUGE-1) 
		&
		(b) Summ. (Runtime) 
		&
		(c) Machine translation
		&
		(d) Dialogue generation

	\end{tabular}
 	\vspace*{-3mm}
	\caption{Comparison with database partitioning strategies for language generation tasks.}\label{fig:partition}
	\vspace*{-3mm}
\end{figure*}
We discuss our proposal with a motivating experiment illustrated in Figure~\ref{fig:partition}. We investigate various strategies for partitioning a database (elaborated in Section~\ref{sec:partition}), and perform RAG with varying the number of partitions for three generation tasks: summarization, translation, and dialogue generation, where we explore all partitions for the retrieval, and the best result (assessed based on a development set) across different partitions is reported. We observe that the optimal performance is typically not achieved through retrieval based on the entire database (\#Partitions $=1$). This observation inspires us to investigate a novel RAG setting with multiple partitions. To achieve this, the task should address three significant challenges, summarized below. (1) Determining a strategy for partitioning a database and the number of partitions. (2) Developing a method for selecting a suitable partition for a given input query to discover effective memories. (3) Enhancing memory quality, including inherent issues such as hallucination, or irrelevant context, which can impact the grounding of LLM generation.

Building upon the aforementioned discussion, we introduce a new solution called \texttt{M-RAG}, designed to facilitate RAG across multiple partitions of a database. \texttt{M-RAG} addresses all of the three challenges. For (1), we draw insights from the literature on vector database management~\cite{pan2023survey,han2023comprehensive} and assess various strategies, namely Randomization~\cite{indyk1998approximate}, Clustering~\cite{jegou2010product}, Indexing~\cite{malkov2014approximate,malkov2018efficient}, and Category~\cite{gollapudi2023filtered}, through empirical studies. The effectiveness of these strategies, along with the corresponding number of partitions, is evaluated across different generative tasks on a development set in our experiments. For (2), with multiple partitions at play, we formulate partition selection as a multi-armed bandit problem~\cite{slivkins2019introduction}. In this context, an agent, denoted as Agent-S, iteratively selects one among several partitions. The characteristics of each partition are only partially known at the time of selection, and Agent-S gains a better understanding over time by maximizing cumulative rewards in the environment. To optimize the decision policy, we leverage reinforcement learning with a carefully designed Markov Decision Process (MDP). For (3), after selecting a partition and obtaining memories for generation, we introduce another agent, denoted as Agent-R. This agent generates a pool of candidate memories iteratively through the use of LLMs. Once a candidate is selected, Agent-R evaluates its quality by demonstrating it to generate a hypothesis. The identification of a high-quality hypothesis determined by a specific performance metric, triggers a boosting process, where it signals the exploration and replacement of the previous memory with a superior one, and continues the process. Further, we integrate the efforts of Agent-S and Agent-R through multi-agent reinforcement learning. With a shared objective of enhancing text generation for a given input query, they are jointly optimized through end-to-end training.

Our contributions can be summarized as follows: (1) we propose a multiple partition paradigm for RAG, aiming to facilitate fine-grained retrieval and concentrate on pivotal memories to enhance overall performance. In addition, the utilization of multiple partitions benefits other aspects of RAG, including facilitating the construction and maintenance of indices, protecting user privacy data within specific partitions, and supporting distributed parallel processing across different partitions. (2) We introduce \texttt{M-RAG}, a new solution based on multi-agent reinforcement learning that tackles the three challenges in executing RAG across multiple partitions. We show that the training objective of \texttt{M-RAG} is well aligned with that of text generation tasks. (3) We conduct extensive experiments on \emph{seven} datasets for \emph{three} generation tasks on \emph{three} distinct language model architectures, including a recent Mixture of Experts (MoE) architecture~\cite{DBLP:journals/corr/abs-2401-04088}. The results demonstrate the effectiveness of \texttt{M-RAG} across diverse RAG baselines. In comparison to the best baseline approach, \texttt{M-RAG} exhibits improvements of 11\%, 8\%, and 12\% for text summarization, machine translation, and dialogue generation tasks, respectively. 
\section{Related Work}
\label{sec:related}

\textbf{Retrieval-Augmented Generation.} We review the literature of Retrieval-Augmented Generation (RAG) in terms of (1) Naive RAG, (2) Advanced RAG, and (3) Modular RAG. 
For (1), Naive RAG follows a standard process including indexing, retrieval, and generation~\cite{DBLP:journals/corr/abs-2305-14283}. However, its quality faces significant challenges such as low precision, hallucination, and redundancy during the process.
For (2), Advanced RAG is further developed to overcome the shortcomings of Naive RAG. Specifically, during the indexing stage, the objective is to enhance the quality of the indexed content by optimizing data embedding~\cite{DBLP:conf/acl/LiLXY00023}. During the retrieval stage, the focus is on identifying the appropriate context by calculating the similarity between the query and chunks, where the techniques involve fine-tuning embedding models~\cite{cocktail}, or learning dynamic embeddings for different context~\cite{DBLP:conf/emnlp/KarpukhinOMLWEC20}. During the generation stage, it merges the retrieved context with the query as an input into large language models (LLMs), where it addresses challenges posed by context window limits with re-ranking the most relevant content~\cite{DBLP:conf/emnlp/JiangWLYQ23, DBLP:conf/emnlp/Zhuang0KZ23}, or compressing prompts~\cite{DBLP:conf/cvpr/LitmanATLMM20, DBLP:journals/corr/abs-2310-04408}. In addition, Self-RAG~\cite{DBLP:journals/corr/abs-2310-11511} is proposed to identify whether retrieval is necessary, or the retrieved context is relevant, which helps language models to produce meaningful generation~\cite{DBLP:journals/corr/abs-2310-11511}.
%
For (3), Modular RAG diverges from the traditional Naive RAG structure by incorporating external modules to further enhance the performance, including search module~\cite{DBLP:journals/corr/abs-2308-11761}, memory module~\cite{DBLP:conf/acl/WangXFLSX0022, cheng2023lift}, tuning module~\cite{DBLP:journals/corr/abs-2310-01352}, and task adapter~\cite{DBLP:conf/emnlp/ChengHBZLW0WDZ23, DBLP:conf/iclr/DaiZMLNLBGHC23}. Specifically, Selfmem~\cite{cheng2023lift} incorporates a retrieval-enhanced generator to iteratively create a memory pool, it then trains a selector to choose one of the memories from the pool to generate responses.
%
The work~\cite{DBLP:journals/corr/abs-2312-10997} provides a comprehensive survey of RAG for LLMs. Our work differs from existing RAG studies in two aspects. First, we introduce a multiple partition setting, where each partition serves as a fundamental entity for retrieval, rather than retrieving from the entire database. Second, we introduce an \texttt{M-RAG} framework built upon multi-agent reinforcement learning, which tackles three distinct challenges posed by this novel setting.

\noindent\textbf{Reinforcement Learning for LLMs.} Recently, reinforcement learning has seen broad applications across a variety of language-related tasks for Large Language Models (LLMs). This includes tasks such as text summarization~\cite{wu2021recursively}, machine translation~\cite{kreutzer2018can}, dialogue systems~\cite{jaques2019way,yi2019towards}, semantic parsing~\cite{lawrence2018improving}, and review generation~\cite{cho2018towards}. For example, WebGPT~\cite{DBLP:journals/corr/abs-2112-09332} incorporates a reinforcement learning framework to autonomously train the GPT-3 model using a search engine during the text generation process. Further, InstructGPT~\cite{ouyang2022training} collects a dataset containing desired model outputs provided by human labelers. Subsequently, it employs Reinforcement Learning from Human Feedback (RLHF) to fine-tune GPT-3~\cite{brown2020language}. In addition, R3~\cite{DBLP:journals/corr/abs-2305-14283} introduces a Rewrite-Retrieve-Read process, where the LLM performance serves as a reinforcement learning incentive for a rewriting module. This approach empowers the rewriter to enhance retrieval queries, consequently improving the reader's performance in downstream tasks. MMQS~\cite{DBLP:conf/www/WangGS24} introduces a new multimodal question suggestion task with a multi-agent version of RLHF.
In this work, we propose a novel multi-agent reinforcement learning framework utilizing two agents to collaboratively optimize text generation tasks. To our best knowledge, this is the first of its kind.

\noindent\textbf{Multi-source Knowledge-grounded Dialogue System (MKDS).} We review the literature on MKDS~\cite{DBLP:conf/emnlp/WuLWZZW21,DBLP:conf/acl/WuLZW22}, and highlight differences with our \texttt{M-RAG} regarding (1) datasets, (2) solutions, and (3) tasks. For (1), MKDS uses multi-source heterogeneous data (plain text, tables, knowledge graphs), each contributing uniquely to dialogue generation. \texttt{M-RAG} uses a single-source homogeneous dataset, initially vectorized and indexed for RAG retrieval. We explore partitioning strategies to create multiple homogeneous partitions for effective retrieval. For (2), MKDS employs an encoder-decoder framework with varied attention weights for different knowledge sources, trained with a small dialogue model like MSKE-Dialog (59.14M parameters)~\cite{DBLP:conf/emnlp/WuLWZZW21}. \texttt{M-RAG} uses a Retrieval-then-Generation approach with two RL agents (Agent-S and Agent-R) focusing on retrieval and generation, respectively. For (3), \texttt{M-RAG} leverages LLMs for diverse language generation tasks, including text summarization, machine translation, and dialogue generation, unlike MKDS's specific focus on dialogue generation~\cite{DBLP:conf/emnlp/WuLWZZW21,DBLP:conf/acl/WuLZW22}.
\section{Methodology}
\label{sec:method}

\begin{figure*}[t]
    \centering
    \includegraphics[width=\linewidth]{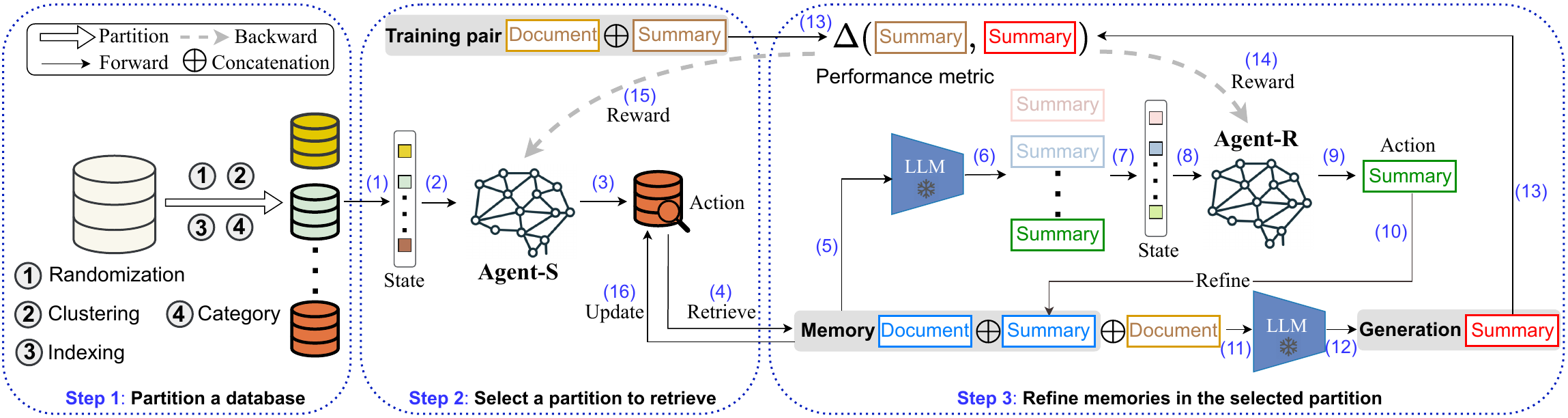}
    \vspace*{-4mm}
    \caption{Illustration of \texttt{M-RAG} training in a summarization task: The \texttt{M-RAG} initiates training with multiple partitions (Section~\ref{sec:partition}), it then selects a partition to perform retrieval via Agent-S (Section~\ref{sec:agent-s}), and refines the memories within the selected partition via Agent-R (Section~\ref{sec:agent-r}). Both agents are collaboratively trained to enhance generation capabilities through multi-agent reinforcement learning (Section~\ref{sec:mrag}). For inference, it includes elements (1), (2), (3), (4), (11), and (12).}
    \label{fig:overall}
    \vspace{-4mm}
\end{figure*}

A task involving \texttt{M-RAG} can be formulated below. Given a database $\mathbb{D}=\{(x_i,y_i)\}_{i=1}^{|\mathbb{D}|}$ for a language generation task (e.g., summarization), where each pair $(x,y)$ represents a document and its corresponding summary stored in $\mathbb{D}$. The \texttt{M-RAG} initiates the process by partitioning $\mathbb{D}$ into multiple partitions. This can be achieved through methods like clustering or by leveraging inherent category labels in the data. The resulting partitions are denoted as $\mathbb{D} = \{D_m\}_{m=1}^{|M|}$, where each $D_m$ $(1 \le m \le M)$ supports an independent RAG process (Section~\ref{sec:partition}).
The \texttt{M-RAG} framework comprises both training and inference processes, as outlined in Algorithm~\ref{alg:mrag}. For training, Agent-S learns to select a specific $D_m$ for an input text pair (Section~\ref{sec:agent-s}). Subsequently, Agent-R refines the retrieved memories, represented as $(\tilde{x}, \tilde{y}) \in D_m$, within the selected partition $D_m$ (Section~\ref{sec:agent-r}). Finally, the two agents are collaboratively trained with multi-agent reinforcement learning (see Section~\ref{sec:mrag}). Figure~\ref{fig:overall} illustrates the training process of \texttt{M-RAG}. For inference, the refined $\mathbb{D}$ is utilized to support a LLM in generating hypotheses, where a $D_m$ is selected by the trained Agent-S.

\subsection{Discussion on Partitioning a Database}
\label{sec:partition}
As \texttt{M-RAG} relies on multiple partitions for RAG operations, we investigate various strategies to partition an external database (typically the training corpus). The results of these strategies are then validated through empirical studies. We review the literature, including recent vector database surveys~\cite{pan2023survey,han2023comprehensive}, and identify the following strategies: namely (1) Randomization~\cite{indyk1998approximate}, (2) Clustering~\cite{jegou2010product}, (3) Indexing~\cite{malkov2014approximate,malkov2018efficient} and (4) Category~\cite{gollapudi2023filtered}. 
Specifically, for (1), it targets the utilization of probability amplification techniques, such as locality-sensitive hashing (LSH), to hash similar items (data vectors) into the same bucket with a high probability. For (2), it involves clustering data vectors using K-means, where this clustering concept is widely applied in Inverted File Index (IVF) for tasks like Approximate k-Nearest Neighbor (AKNN) search. For (3), navigable graph indexes, such as HNSW~\cite{malkov2018efficient} or NSW~\cite{malkov2014approximate}, are designed to facilitate easy traversal of different regions within a vector database. To achieve effective partitions, we employ graph partitioning with spectral clustering on a navigable graph. For (4), it involves assigning data vectors to partitions based on their respective categories. For example, in the DailyDialog dataset~\cite{DBLP:conf/ijcnlp/LiSSLCN17}, which includes 7 emotion categories (e.g., joy, anger) and 10 topic categories (e.g., work, health), vectors are partitioned according to their category labels. We note that a single vector may be assigned to multiple partitions, due to the characteristics of the dataset, where a dialogue spans multiple categories.

In Figure~\ref{fig:partition}, we perform experiments on a development set, manipulating the number of partitions wrt the 4 strategies across three language generation tasks (summarization, translation, and dialogue generation). The results demonstrate the effectiveness of the strategies, and we conclude the selected strategies with the number of partitions as follows. We choose Indexing (4 partitions), Randomization (3 partitions), and Category (10 partitions) for the summarization, translation, and dialogue generation tasks, respectively. In addition, as shown in Figure~\ref{fig:partition} (a) and (b), we observe that both Top-1 and Top-3 retrieval methods exhibit comparable performance. For enhanced efficiency, we default to Top-1 retrieval in the rest of the paper.

\subsection{Agent-S: Selecting a Database Partition}
\label{sec:agent-s}
During the training process of an Agent-S to select a partition from $\mathbb{D}$, the environment is naturally modeled as a bandit setting. In this context, when a random partition is selected, the language model generates a response for the query with feedback (typically based on a specific performance metric), and concludes the episode. The selection process can be formulated as a Markov Decision Process (MDP), involving states, actions, and rewards. 

\noindent\textbf{States.} Given a training pair $(x, y)$ and a set of database partitions $\mathbb{D} = \{D_m\}_{m=1}^{|M|}$, the state $s^{(S)}$ is defined by assessing the semantic relevance, typically quantified by measures such as cosine similarity $\text{sim}(\cdot,\cdot)$, between the input $(x, y)$ and the stored memories $(\tilde{x}, \tilde{y})$ within each $D_m$.
\begin{align}
    \label{eq:state_s}
    s^{(S)} = \{ \max\limits_{(\tilde{x},\tilde{y}) \in D_m} \text{sim}(\sigma(\tilde{x} \oplus \tilde{y}), \sigma(x \oplus y)) \}_{m=1}^{|M|},
\end{align}
where $\oplus$ denotes the concatenation operation, and $\sigma(\cdot)$ denotes an embedded model utilized to obtain text representations, such as the CPT-Text~\cite{neelakantan2022text}. We consider the Top-1 retrieved memories to construct the state. 

\noindent\textbf{Actions.} Let $a^{(S)}$ represent an action undertaken by Agent-S. The design of actions corresponds to that of the state $s^{(S)}$. Specifically, the actions are defined as follows:
\begin{align}
    \label{eq:action_s}
    a^{(S)} = m \ (1 \le m \le M),
\end{align}
where action $a^{(S)} = m$ means to select the $D_m$ for subsequent the generation task.

\noindent\textbf{Rewards.} The reward is denoted by $r^{(S)}$. When the action $a^{(S)}$ involves exploring a partition, the reward cannot be immediately observed, as no response has been received for the query $x$. However, when the action involves selecting a partition for Agent-R to refine the memories within the partition, the stored response $\tilde{y}$ is updated, and some reward signal can be obtained (for example, by measuring the difference between the results on the original memory and that on the refined memory). 
Therefore, we make Agent-S and Agent-R are trained with multi-agent reinforcement learning, since they cooperate towards the same objective of learning a policy that produces a response (hypothesis) as similar as possible to the reference $y$ for the $x$.


\subsection{Agent-R: Refining Memories in the Selected Partition}
\label{sec:agent-r} 

Next, we formulate the task of refining the retrieved memories carried out by Agent-R within a selected partition. To accomplish this, Agent-R explores potential responses denoted by $\hat{y}$ through LLMs for the retrieved $\tilde{x}$, and generates a candidate pool $\mathbb{C} = \{ \hat{y}_k \leftarrow \text{LLM}(\tilde{x}) \}_{k=1}^{|K|}$ for selection, where $K$ denotes the number of candidates. Upon selecting a candidate, Agent-R evaluates its quality by demonstrating the new memory $(\tilde{x}, \hat{y}_k)$ to generate a hypothesis $h \leftarrow \text{LLM}(x \oplus (\tilde{x}, \hat{y}_k))$. In summary, a high-quality hypothesis $h$ benefits from superior memory, which can be then refined through the produced hypothesis for subsequent selections. Consequently, Agent-R iterates in a boosting process optimized via reinforcement learning, where the states, actions, and rewards are detailed below.

\noindent\textbf{States.} The state $s^{(R)}$ is defined to assess the semantic relevance between the produced hypothesis $h$ and the selected $\hat{y}_k$ from the pool $\mathbb{C}$. The rationale is to identify a memory that closely resembles the hypothesis, which aligns with the human intuition that a superior demonstration sample often leads to better generation results, that is
\begin{align}
    \label{eq:state_r}
    s^{(R)} = \{\text{sim}(\sigma(h), \sigma(\hat{y}_k)) \}_{k=1}^{|K|},
\end{align}
where $\sigma(\cdot)$ denotes an embedded model, and $K$ governs the constructed state space.

\noindent\textbf{Actions.} Let $a^{(R)}$ represent an action taken by Agent-R. The design is consistent with the state $s^{(R)}$, which involves selecting a candidate memory from the pool, that is
\begin{align}
    \label{eq:action_r}
    a^{(R)} = k \ (1 \le k \le K).
\end{align}

\noindent\textbf{Rewards.} We denote the reward of Agent-R as $r_t^{(R)}$, which corresponds to the transition from the current state $\mathbf{s}_t^{(R)}$ to the next state $\mathbf{s}_{t+1}^{(R)}$ after taking action $a_t^{(R)}$. Specifically, when a memory $(\tilde{x}, \hat{y}_k)$ is updated, the hypothesis changes from $h$ to $h'$ accordingly. We remark that the best hypothesis (denoted as $h'$) identified at state $s^{(R)}$ is maintained according to a specific metric $\Delta(\cdot,\cdot)$ (e.g., ROUGE for text summarization, BLEU for machine translation, BLEU and Distinct for dialogue generation), and the reward is defined as:

\SetKwInOut{KwIn}{Require}
\begin{algorithm}[t!]
    \small
    \caption{The \texttt{M-RAG} Framework}
    \label{alg:mrag}
	\KwIn{
        a database $\mathbb{D}$; a frozen $\text{LLM}(\cdot)$}
        obtain $\mathbb{D} = \{D_m\}_{m=1}^{|M|}$ via a partitioning strategy\\
        initialize Ag-S $\pi_{\theta}(a^{(S)}|s^{(S)})$, Ag-R $\pi_{\phi}(a^{(R)}|s^{(R)})$ \\
        \While {not converged on a validation set} { 
        sample a text pair $(x,y)$ from the training set \\
        construct $s_1^{(S)}$ with $(x,y)$ on $\mathbb{D}$ by Eq~\ref{eq:state_s} \\ 
        \For{$i=1,2,...$} { 
            sample $m = a_i^{(S)} \sim \pi_{\theta}(a|s_i^{(S)})$ \\
            $r_i^{(S)} \leftarrow 0$\\
            $h \leftarrow \text{LLM}(x \oplus (\tilde{x}, \tilde{y}) \in D_m)$\\
            construct $s_1^{(R)}$ with $h$ on $\mathbb{C} = \{ \hat{y}_k \leftarrow \text{LLM}(\tilde{x}) \}_{k=1}^{|K|}$
            by Eq~\ref{eq:state_r} \\ 
            
            \For{$j=1,2,...$}{
            sample $k = a_j^{(R)} \sim \pi_{\phi}(a|s_j^{(R)})$ \\    
            $h' \leftarrow \text{LLM}(x \oplus (\tilde{x}, \hat{y}_k))$\\
            
            \uIf{$\Delta(h', y) > \Delta(h, y)$}{
             $r_j^{(R)} \leftarrow \Delta(h', y)-\Delta(h, y)$\\ $D_m.{\tilde{y}} \leftarrow \hat{y}_k$, $h \leftarrow h'$\\
             }
             \uElse{
                $r_j^{(R)} \leftarrow 0$
             }
            construct $s_{j+1}^{(R)}$ with $h$ on a new $\mathbb{C}$\\
            $r_i^{(S)} \leftarrow r_i^{(S)} + r_j^{(R)}$\\
            }
        construct $s_{i+1}^{(S)}$ by updating $(\tilde{x}, \tilde{y})$ and $(x,y)$\\
        optimize $\pi_{\theta}$ and $\pi_{\phi}$ via DQN 
        }
    }
generate final hypotheses via $\text{LLM}(\cdot)$ on $\mathbb{D}$ (where the trained Ag-S selects a partition)  
\end{algorithm}

\begin{align}
\label{eq:reward_r}
r^{(R)} = \Delta(h', y)-\Delta(h, y),
\end{align}
where $y$ denotes the reference result. In this reward definition, we observe that the objective of the Markov Decision Process (MDP), which aims to maximize cumulative rewards, aligns with Agent-R's goal of discovering the best hypothesis among the memories. To illustrate, we consider the process through a sequence of states: $s^{(R)}_1, s^{(R)}_2, ..., s^{(R)}_N$, concluding at $s^{(R)}_N$. The rewards received at these states, except for the termination state, can be denoted as $r^{(R)}_1, r^{(R)}_2, ..., r^{(R)}_{N-1}$. When future rewards are not discounted, we have:
\begin{equation}
    \begin{aligned}
    \label{eq:rwd}
    \sum_{t=2}^{N}r^{(R)}_{t-1} &=\sum_{t=2}^{N}(\Delta(h_{t}, y) - \Delta(h_{t-1}, y)) \\
    &= \Delta(h_N, y) - \Delta(h_1, y),
\end{aligned}
\end{equation}
where $\Delta(h_N, y)$ corresponds to the highest hypothesis value found throughout the entire iteration, and $\Delta(h_1, y)$ represents an initial value that remains constant. Therefore, maximizing cumulative rewards is equivalent to maximizing the discovered hypothesis value. Finally, the cumulative reward is shared with Agent-S to align with the training objective, that is
\begin{align}
\label{eq:reward_s}
r^{(S)} = \Delta(h_N, y) - \Delta(h_1, y).
\end{align}

\subsection{The \texttt{M-RAG} Framework}
\label{sec:mrag}
\noindent\textbf{Policy Learning via DQN.} In a MDP, the primary challenge lies in determining an optimal policy that guides an agent to select actions at states, with the aim of maximizing cumulative rewards. Given that the states within our MDPs are continuous, we employ Deep Q-Networks (DQN) with replay memory~\cite{mnih2013playing} to learn the policy, denoted as $\pi_{\theta}(a^{(S)}|s^{(S)})$ for Agent-S (resp. $\pi_{\phi}(a^{(R)}|s^{(R)})$ for Agent-R). The policy samples an action $a^{(S)}$ (resp. $a^{(R)}$) at a given state $s^{(S)}$ (resp. $s^{(R)}$) via DQN, with parameters denoted by $\theta$ (resp. $\phi$).

\noindent\textbf{Combining Agent-S and Agent-R.} We present the \texttt{M-RAG} framework in Algorithm~\ref{alg:mrag}, which combines the functionalities of Agent-S and Agent-R on multiple partitions (line 1). The algorithm comprises two main phases: training and inference. During the training phase (lines 2-22), we randomly sample text pairs from the training set (line 4). For each pair, we generate episodes to iteratively train Agent-S and Agent-R, with the MDPs outlined in (lines 6-21) and (lines 11-20), respectively. Experiences of $(s^{(S)}_t, a^{(S)}_t, r^{(S)}_t, s^{(S)}_{t+1})$ and $(s^{(R)}_t, a^{(R)}_t, r^{(R)}_t, s^{(R)}_{t+1})$ are stored during the iteration, and a minibatch is sampled to optimize the two agents via DQN (line 22). 

During the inference phase (line 23), final hypotheses are generated via a LLM based on the refined $\mathbb{D}$, where a partition is selected by the trained Agent-S, and the $\tilde{y}$ and $y$ (unknown during inference) are omitted to construct the state by Eq~\ref{eq:state_s}.

\noindent\textbf{Time Complexity.} We discuss the complexity of \texttt{M-RAG} compared to a Naive RAG setup introduced in Section~\ref{sec:related} in terms of the three steps: (1) indexing, (2) retrieval, and (3) generation as shown in Figure~\ref{fig:overall}. In terms of inference, involving (1) and (2), it is worth noting that the \texttt{M-RAG} exhibits a complexity comparable to that of a Naive RAG setup, with the additional complexity (3) only being involved during training.

For (1), the complexity associated with constructing multiple partitions (e.g., using the HNSW index structure) is represented as $O(M \cdot N \log N)$, where $M$ indicates the number of partitions and $N$ indicates the maximum number of memories within a partition. This approach proves to be faster compared to a Naive RAG setup, which organizes all data within a single index structure with a construction complexity of $O(N' \log N')$, where $N'$ represents the total number of memories in the database.

For (2), the complexity of Agent-S is approximately $O(M \cdot \log N)$, where an AKNN search is performed within each partition, incurring a cost of $O(M \cdot \log N)$ with HNSW. Additionally, sampling actions via Agent-S requires $O(1)$ complexity, owing to its lightweight neural network architecture. In contrast, for the Naive RAG setup, conducting the AKNN search within the entire database costs $O(\log N')$, which is marginally faster than the \texttt{M-RAG} setup.

For (3), the complexity of Agent-R is roughly $O(C \cdot E^2)$, where $E$ tokens are generated via a LLM based on the transformer attention mechanism, and $C$ represents the number of its MDP iterations. This component predominantly influences the overall training complexity. In contrast, for a Naive RAG setup, it runs only once during the inference procedure to produce the generation outcomes, with a complexity of approximately $O(E^2)$.

\section{Experiments}

\subsection{Experimental Setup}
\label{sec:setup}

\noindent \textbf{Datasets.}
By following~\cite{cheng2023lift}, we conduct experiments on seven datasets for three generation tasks: (1) text summarization (XSum~\citealp{DBLP:conf/emnlp/NarayanCL18} and BigPatent~\citealp {DBLP:conf/acl/SharmaLW19}), (2) machine translation (JRC-Acquis~\citealp{DBLP:conf/lrec/SteinbergerPWIE06} with Es$\rightarrow$En, En$\rightarrow$Es, De$\rightarrow$En, and En$\rightarrow$De), and (3) dialogue generation (DailyDialog~\citealp{DBLP:conf/ijcnlp/LiSSLCN17}). Specifically, XSum comprises single-document summaries for highly abstractive articles sourced from BBC news. BigPatent comprises 1.3 million records of U.S. patent documents accompanied by human-written abstractive summaries. JRC-Acquis serves as a collection of parallel legislative texts of European Union Law, commonly employed as a benchmark in machine translation tasks. DailyDialog comprises multi-turn dialogues centered around daily life topics. The detailed statistics for these datasets are available in~\cite{cheng2023lift}.

\noindent \textbf{Baselines.} We carefully review the literature including a recent survey paper~\cite{DBLP:journals/corr/abs-2312-10997}, and identify the following RAGs, namely Naive RAG~\cite{DBLP:journals/corr/abs-2305-14283}, Self-RAG~\cite{DBLP:journals/corr/abs-2310-11511}, and Selfmem~\cite{cheng2023lift}, which correspond to three kinds of RAG techniques as described in Section~\ref{sec:related}. In addition, we incorporate the RAGs into three typical language model architectures, namely Mixtral 8$\times$7B~\cite{DBLP:journals/corr/abs-2401-04088}, Llama 2 13B~\cite{DBLP:journals/corr/abs-2307-09288}, Phi-2 2.7B~\cite{phi2}, Gemma 7B~\cite{DBLP:journals/corr/abs-2403-08295}, and Mistral 7B~\cite{DBLP:journals/corr/abs-2310-06825} for the evaluation.

\noindent \textbf{Evaluation Metrics.}
We evaluate the effectiveness of \texttt{M-RAG} in terms of the three generation tasks by following~\cite{cheng2023lift}. (1) For summarization, ROUGE (R-1/2/L)~\cite{lin-2004-rouge} is used. (2) For machine translation, BLEU~\cite{DBLP:conf/wmt/Post18} is used. (3) For dialogue generation, BLEU (B-1/2) and Distinct (D-1/2)~\cite{DBLP:conf/naacl/LiGBGD16, DBLP:conf/nips/LiXYSZL21} are used. Overall, a higher evaluation metric (i.e., ROUGE, BLEU, Distinct) indicates a better result. We remark that all results are statistically significant, as confirmed by a t-test with $p<0.05$.

\noindent \textbf{Implementation Details.} We implement \texttt{M-RAG} and adapt other baselines using Python 3.7 and LlamaIndex. The database partitioning strategies for Randomization~\footnote{https://pypi.org/project/graph-partition/} and Indexing~\footnote{https://pypi.org/project/LocalitySensitiveHashing/} utilize existing libraries. The Agent-S (resp. Agent-R) is instantiated through a two-layered feedforward neural network. The first layer consists of 25 neurons using the tanh activation function, and the second layer comprises $M$ (resp. $K$) neurons corresponding to the action space with a linear activation function. The hyperparameters $M$ and $K$ are empirically set to 4 and 3, respectively. Some of the built-in RL codes can be found in the GitHub repositories referenced in~\cite{DBLP:journals/corr/abs-2311-11204, DBLP:conf/kdd/WangLCZ21}. During training, we randomly sample 10\% of text pairs from the training set, while the remaining data is utilized for constructing the database with multiple partitions. The MDP iterations are determined by performance evaluation on a validation set. Evaluation metrics, such as ROUGE, BLEU, and Distinct, are obtained from~\cite{cheng2023lift}. The language models with 4-bit quantization, including Mixtral 8$\times$7B, Llama 2 13B, Phi-2 2.7B, Gemma 7B, and Mistral 7B, are available for download via the link~\footnote{https://huggingface.co/TheBloke}. To boost training efficiency, we cache the QA pairs generated by the LLMs during training.
\subsection{Experimental Results}
\label{sec:result}

\begin{table*}[t]
\caption{Text summarization.}
\centering
\vspace{-4mm}
\label{tab:summarization}
\begin{tabular}{l|c|c|ccc|ccc}
\hline
\multicolumn{2}{l|}{\multirow{2}{*}{LLM}}  & \multirow{2}{*}{RAG} & \multicolumn{3}{c|}{XSum} & \multicolumn{3}{c}{BigPatent} \\ \cline{4-9} 
\multicolumn{2}{l|}{}                      &                      & R-1     & R-2    & R-L    & R-1      & R-2      & R-L     \\ \hline
\multicolumn{2}{l|}{Mixtral 8 $\times$ 7B} & None                &    25.40      &   6.39     &   18.30     &    47.41      &    16.63    &    25.14    \\
\multicolumn{2}{l|}{Mixtral 8 $\times$ 7B} & Naive              & 43.82 & 22.07 & 37.44        & {\Es 60.11}         & {\Es 38.33}         & {\Es 43.44}        \\
\multicolumn{2}{l|}{Mixtral 8 $\times$ 7B} & Selfmem              &   44.67      &  22.38     &   37.86     &  {\Es 64.12}        &  {\Es 39.21}        &   {\Es 46.21}      \\
\multicolumn{2}{l|}{Mixtral 8 $\times$ 7B} & Self-RAG             &    44.01     &  22.26      &    37.51    &  {\Es 63.59}        &  {\Es 38.65}        &   {\Es 45.25}      \\ \hline
\multicolumn{2}{l|}{Mixtral 8 $\times$ 7B} & \texttt{M-RAG}                & \textbf{48.13}    & \textbf{24.66}  & \textbf{39.43}        &  {\Es \textbf{71.34}}        &  {\Es \textbf{42.24}}        &   {\Es \textbf{47.22}}      \\
\multicolumn{2}{l|}{Llama 2 13B}           & \texttt{M-RAG}               &    37.18     &   18.02     &    26.44    &    {\Es 60.31}      &  {\Es 37.33}        &    {\Es 33.47}     \\
\multicolumn{2}{l|}{Phi-2 2.7B}            & \texttt{M-RAG}                &    30.70     &  11.57      &    26.20    &     31.25     &    14.72      &    18.98     \\ \hline
\end{tabular}
\vspace{-3mm}
\end{table*}

\begin{table*}[t]
\caption{Machine translation.}
\centering
\vspace{-4mm}
\label{tab:translation}
\begin{tabular}{l|c|c|cc|cc|cc|cc}
\hline
\multicolumn{2}{l|}{\multirow{2}{*}{LLM}} & \multirow{2}{*}{RAG} & \multicolumn{2}{c|}{Es$\rightarrow$En} & \multicolumn{2}{c|}{En$\rightarrow$Es} & \multicolumn{2}{c|}{De$\rightarrow$En} & \multicolumn{2}{c}{En$\rightarrow$De} \\ \cline{4-11} 
\multicolumn{2}{l|}{}                     &                      & Dev               & Test               & Dev               & Test               & Dev               & Test               & Dev               & Test              \\ \hline
\multicolumn{2}{l|}{Mixtral 8 $\times$ 7B} & None                 &        34.34           &          34.81          &            32.60       &          28.32         &             43.75      &        44.09           &              43.78     &       42.24           \\
\multicolumn{2}{l|}{Mixtral 8 $\times$ 7B} & Naive              &         36.64          &         36.22            &        33.18           &          30.70          &            47.84       &        46.77             &               45.83    &        44.23           \\
\multicolumn{2}{l|}{Mixtral 8 $\times$ 7B} & Selfmem              &        37.65           &        37.11            &        34.12           &         31.86           &            48.08       &         47.31           &   51.38                &         49.81          \\
\multicolumn{2}{l|}{Mixtral 8 $\times$ 7B} & Self-RAG             &         37.17          &           36.82         &         33.80          &         31.61           &            47.99       &         47.27           &   50.10                &         48.75          \\ \hline
\multicolumn{2}{l|}{Mixtral 8 $\times$ 7B} & \texttt{M-RAG}                &         \textbf{39.11}          &    \textbf{39.98}                &       \textbf{35.18}            &  \textbf{32.70}              &  \textbf{49.16}            &      \textbf{48.15}              &       \textbf{53.76}            &      \textbf{50.75}            \\
\multicolumn{2}{l|}{Llama 2 13B}          & \texttt{M-RAG}                &      30.41             &      30.03              &        26.40           &       22.03             &      41.10             &        42.22            &       45.98            &        42.58           \\
\multicolumn{2}{l|}{Phi-2 2.7B}           & \texttt{M-RAG}                &    22.83               &       24.22             &       17.64            &         16.60          &         34.21          &          34.71          &      40.01             &        37.08           \\ \hline
\end{tabular}
\vspace{-5mm}
\end{table*}

\noindent \textbf{(1) Effectiveness evaluation (partitioning strategies).} We conduct experiments to evaluate various partitioning strategies across text summarization (XSum), machine translation (Es$\rightarrow$En), and dialogue generation (DailyDialog) tasks with Mixtral 8 $\times$ 7B. The best results, based on a development set across different partitions, are reported. As shown in Figure~\ref{fig:partition}, we observe that retrieval based on the entire database generally fails to achieve optimal performance. Moreover, the performance slightly decreases as the number of partitions increases. This is attributed to the AKNN search, where a smaller partition size recalls more similar memories, which may not align well with the LLM preferences and impede the focus on crucial memories. Additionally, we observe that the RAG with Top-1 retrieval exhibits faster runtime compared to the Top-3 due to a shorter input length for the LLM, while maintaining comparable performance.

\begin{table}[t]
\caption{Dialogue generation.}
\setlength{\tabcolsep}{1.8pt}
\vspace{-4mm}
\label{tab:dialogue}
\begin{tabular}{l|c|c|cccc}
\hline
\multicolumn{2}{l|}{\multirow{2}{*}{LLM}}  & \multirow{2}{*}{RAG} & \multicolumn{4}{c}{DailyDialog} \\ \cline{4-7} 
\multicolumn{2}{l|}{}                      &                      & B-1    & B-2    & D-1   & D-2   \\ \hline
\multicolumn{2}{l|}{Mix. 8 $\times$ 7B} & None                 &    15.52    &  7.05      &  61.49      &  89.51    \\
\multicolumn{2}{l|}{Mix. 8 $\times$ 7B} & Naive              &    37.44    &  29.16      &  89.42     &  92.55     \\
\multicolumn{2}{l|}{Mix. 8 $\times$ 7B} & Selfmem              &  {\Es 38.16}      & {\Es 29.92}       &  {\Es 89.23}     &  {\Es 95.23}     \\
\multicolumn{2}{l|}{Mix. 8 $\times$ 7B} & Self-RAG             &   {\Es  37.76}    &  {\Es  29.79}     &    {\Es 88.24} &  {\Es 95.34}      \\ \hline
\multicolumn{2}{l|}{Mix. 8 $\times$ 7B} & \texttt{M-RAG}             &  \textbf{42.61}    &   \textbf{32.97}     &  88.82     &   95.74     \\
\multicolumn{2}{l|}{Llama 2 13B}           & \texttt{M-RAG}                &   31.29     &   17.63     &   63.19    &   88.20    \\
\multicolumn{2}{l|}{Phi-2 2.7B}            & \texttt{M-RAG}                &   7.71     &    3.93    &  44.21     &   82.86    \\ \hline
\multicolumn{2}{l|}{Mix. 8 $\times$ 7B} & \texttt{M-RAG(D)}                & {\Es 39.14}       & {\Es 30.98}   & {\Es \textbf{93.14}} & {\Es \textbf{98.34}}  \\ \hline
\end{tabular}
\vspace{-6mm}
\end{table}

\noindent \textbf{(2) Effectiveness evaluation (text summarization).} 
We compare the performance of the \texttt{M-RAG} against alternative RAG methods on three distinct language models: Mixtral 8$\times$7B, Llama 2 13B, and Phi-2 2.7B. The corresponding results are outlined in Table~\ref{tab:summarization}. We observe consistent improvement in language models when utilizing the RAG framework (e.g., Naive) compared to models without RAG (e.g., None). In addition, the recent MoE architecture Mistral 8 $\times$ 7B generally outperforms the typical Llama 2 13B in the summarization task. Specifically, when considering Mistral 8 $\times$ 7B as a base model, the performance of \texttt{M-RAG} outperforms that of other baseline models on both datasets. For example, it achieves better results than the best baseline model Selfmem, by 8\% and 11\% in terms of R-1 on XSum and BigPatent, respectively.

\noindent \textbf{(3) Effectiveness evaluation (machine translation).} 
We further conduct experiments to evaluate the performance of \texttt{M-RAG} for machine translation, and the results are reported in Table~\ref{tab:translation}. We observe that a consistent improvement in the performance of translation tasks with \texttt{M-RAG} across four datasets and three architectures. Notably, it surpasses the Selfmem by 8\% in the Es$\rightarrow$En translation task.

\noindent \textbf{(4) Effectiveness evaluation (dialogue generation).}
As shown in Table~\ref{tab:dialogue}, \texttt{M-RAG} further enhances the language model performance for dialogue generation tasks. It outperforms the Selfmem by 12\% in terms of B-1. Notably, we can also use the Distinct score as the performance metric for optimizing the two agents, denoted by \texttt{M-RAG(D)}, and it results in a more diverse dialogue.

\begin{table*}[]
\caption{Comparing \texttt{M-RAG} on various 7B LLMs.}
\centering
\vspace{-3mm}
\label{tab:7b}
\begin{tabular}{l|c|ccc|c|cc}
\hline
                                                          &                       & \multicolumn{3}{c|}{Summarization}               & Translation (Es$\rightarrow$En) & \multicolumn{2}{c}{Dialogue}    \\ \cline{3-8} 
\multirow{-2}{*}{LLM}                                     & \multirow{-2}{*}{RAG} & R-1            & R-2            & R-L            & BLEU                            & B-1            & B-2            \\ \hline
Gemma 7B                                                  & Selfmem               & 31.38          & 9.97           & 25.07          & 24.61                           & 15.56          & 7.91           \\
Gemma 7B                                                  & M-RAG                 & \textbf{33.81} & \textbf{12.93} & \textbf{27.82} & \textbf{26.92}                  & \textbf{18.15} & \textbf{9.95}  \\ \hline
Mistral 7B & Selfmem               & 35.40          & 12.68          & 27.06          & 26.26                           & 18.28          & 10.05          \\
Mistral 7B                                                & M-RAG                 & \textbf{37.47} & \textbf{13.24} & \textbf{30.49} & \textbf{32.65}                  & \textbf{24.52} & \textbf{11.53} \\ \hline
\end{tabular}
\vspace{-6mm}
\end{table*}

\begin{table}[t]
\caption{Ablation study.}
\setlength{\tabcolsep}{4pt}
\vspace{-3mm}
\label{tab:ablation}
\begin{tabular}{cccc}
\hline
Components                 & R-1 & R-2 & R-L \\ \hline
\texttt{M-RAG}                    & \textbf{48.13}    & \textbf{24.66}  & \textbf{39.43} \\ \hline
w/o Agent-S (single DB) & 44.20 & 22.72 &  37.40  \\
w/o Agent-R (greedy)      & 45.75 & 23.21 & 38.28   \\ \hline
w/o Agent-S and Agent-R       & 43.82 & 22.07 & 37.44   \\ \hline
\end{tabular}
\vspace*{-3mm}
\end{table}

\noindent \textbf{(5) Effectiveness evaluation (results on 7B LLMs).} We increase the number of evaluated LLMs, e.g., comparing 7B models (Gemma 7B and Mistral 7B) to show more results. This comparison aims to assess the performance of \texttt{M-RAG} across the three generation tasks, against the best baseline method Selfmem. The results are presented in Table~\ref{tab:7b}. In general, \texttt{M-RAG} consistently outperforms Selfmem on the 7B models.

\noindent \textbf{(6) Ablation study.} To evaluate the effectiveness of the two agents in \texttt{M-RAG}, we conduct an ablation study on XSum. We remove Agent-S and utilize the entire database for RAG; we replace Agent-R with a greedy rule to select a candidate memory from the pool according to Equation~\ref{eq:state_r}; and we remove both agents, which degrades to the Naive RAG. The results are presented in Table~\ref{tab:ablation}, demonstrating that both agents contribute to performance improvement. Specifically, removing Agent-S results in a significant decline in R-1 from 48.13 to 44.20. This underscores the role of the multiple partition setting in enhancing overall performance. Moreover, removing Agent-R leads to a reduction in R-1 from 48.13 to 45.75. This decline is attributed to the effectiveness of Agent-R in learning memory selection dynamically, as opposed to relying on a fixed rule for decision-making.

\begin{table}[t]
\setlength{\tabcolsep}{2.2pt}
\caption{Impacts of the number of $M$ in Agent-S.}
\vspace{-3mm}
\label{tab:parameter_m}
\begin{tabular}{c|ccccc}
\hline
$M$ & 1 & 2 & 3 & 4 & 5 \\ \hline
R-1       &44.20 &44.53 &46.27 &48.13 &47.21    \\
Index constr. (s) & 299 & 278 & 257 & 246 & 227   \\
Retrieval (s)      & 0.61 & 1.09 & 1.54 & 2.19 & 2.59\\
Generation (s)     & 83.59 & 84.88 & 82.81 & 82.89 & 86.64 \\
\hline
\end{tabular}
\vspace{-6mm}
\end{table}

\begin{table}[t]
\setlength{\tabcolsep}{3.4pt}
\caption{Impacts of the number of $K$ in Agent-R.}
\vspace{-3mm}
\label{tab:parameter_k}
\begin{tabular}{c|ccccc}
\hline
$K$ & 1 & 2 & 3 & 4 & 5 \\ \hline
R-1       &45.81 &46.54 & 48.13 &48.18 &48.25    \\
Pool gen. (s)       & 76 & 191 & 267 & 290 & 359   \\
\hline
\end{tabular}
\vspace{-6mm}
\end{table}

\noindent \textbf{(7) Parameter study (Agent-S state space $M$).} We study the effect of parameter $M$, which controls
the state space of Agent-S and corresponds to the number of partitions. In Table~\ref{tab:parameter_m}, we observe that setting $M = 4$ yields the best effectiveness while maintaining reasonable runtime in terms of index construction, retrieval, and generation. This is consistent with empirical studies illustrated in Figure~\ref{fig:partition} (a). When $M=1$, it reduces to a single database for RAG. As $M$ increases, index construction accelerates on smaller partitions, while retrieval time sightly increases due to the additional time required for constructing states by querying each partition. As expected, the retrieval time is much smaller than the language generation time.

\noindent \textbf{(8) Parameter study (Agent-R state space $K$).} We study the effect of parameter $K$ in Agent-R, representing the state space of Agent-R, to choose one memory from a candidate pool with a size of $K$. In Table~\ref{tab:parameter_k}, we observe a performance improvement as $K$ increases from 1 to 3, and then remains stable. Particularly, when $K=1$, \texttt{M-RAG} exhibits the worst performance, possibly due to the limited exploration of potential memories for generating improved hypotheses. We choose the setting of $K=3$, as it demonstrates effective performance, and runs reasonably fast for generating the pool.

\section{Conclusion and Limitations}
\label{sec:conclusion}

In this paper, we propose a multiple partition paradigm for RAG, which aims to refine retrieval processes and emphasize pivotal memories to improve overall performance. Additionally, we introduce \texttt{M-RAG}, a novel framework with multi-agent reinforcement learning, which addresses key challenges inherent in executing RAG across multiple partitions. The training objective of \texttt{M-RAG} is well aligned with that of text generation tasks, showcasing its potential to enhance system performance explicitly. Through extensive experiments conducted on seven datasets for three language generation tasks, we validate the effectiveness of \texttt{M-RAG}. 

For limitations, we conduct experiments with quantized versions of language models due to computational constraints. However, the observed effectiveness gains are expected to remain consistent across different model sizes and should not significantly impact the overall trends of various RAG methods. Further, although the parameters of the LLMs remain fixed and only the parameters of Agent-S and Agent-R are trained, the training efficiency is limited, as indicated by the training time complexity discussed in Section~\ref{sec:mrag}. This is due to the necessity of querying the LLMs during the training process. In future work, we intend to explore solutions to overcome these limitations.

\bibliography{ref}
\bibliographystyle{acl_natbib}

\appendix \section{Appendix}
\label{sec:appendix}

\begin{table*}[ht]
\setlength{\tabcolsep}{3pt}
\caption{Machine translation with BLEURT and COMET.}
\label{tab:comet}
\begin{tabular}{ll|l|cccc|cccc}
\hline
\multicolumn{2}{l|}{\multirow{2}{*}{LLM}}  & \multirow{2}{*}{RAG}         & \multicolumn{4}{c|}{BLEURT}                                                                                                                  & \multicolumn{4}{c}{COMET}                                                     \\ \cline{4-11} 
\multicolumn{2}{l|}{}                      &                              & \multicolumn{1}{c}{Es$\rightarrow$En} & \multicolumn{1}{c}{En$\rightarrow$Es} & \multicolumn{1}{c}{De$\rightarrow$En} & En$\rightarrow$De & Es$\rightarrow$En & En$\rightarrow$Es & De$\rightarrow$En & En$\rightarrow$De \\ \hline
\multicolumn{2}{l|}{Mixtral 8 $\times$ 7B} & \multicolumn{1}{c|}{Selfmem} & 63.63                                  & 53.26                                  & 59.93                                  & 59.91             & 75.65             & 55.28             & 60.41             & 52.13             \\
\multicolumn{2}{l|}{Mixtral 8 $\times$ 7B} & \multicolumn{1}{c|}{M-RAG}   & \textbf{71.74}                                  & \textbf{63.66}                                  & \textbf{66.77}                                  & \textbf{70.99}             & \textbf{82.66}             & \textbf{80.29}             & \textbf{67.33}             & \textbf{85.14}             \\ \hline
\end{tabular}
\end{table*}

\subsection{Other Evaluation Metrics for Machine Translation} 
\label{asec:metric}
We utilize BLEURT~\footnote{https://huggingface.co/spaces/evaluate-metric/bleurt} (with the checkpoint of BLEURT-20) and COMET~\footnote{https://huggingface.co/spaces/evaluate-metric/comet} (with wmt22-comet-da to obtain features) to evaluate the performance of machine translation, and then compare \texttt{M-RAG} with the best baseline method, Selfmem, on the Mixtral 8 × 7B. The results are reported in Table~\ref{tab:comet}. Overall, we observe that \texttt{M-RAG} consistently outperforms Selfmem across diverse translation datasets, as evidenced by various evaluation metrics.

\subsection{Further Discussion}

\noindent\textbf{Q1. Why applying RAG for summarization or translation?} 

Employing RAG for summarization or translation is based on two key factors: (1) We believe that the two tasks effectively capture the essence of text generation facilitated by LLMs; (2) the widespread adoption of summarization and translation tasks in retrieval-augmented literature~\cite{cheng2023lift, DBLP:conf/aaai/GuWCL18,DBLP:conf/acl/HossainGZ20} provides a standardized and comparable testbed for benchmarking our method. Here, certain text pairs are stored within an external database, such as (document, summary) pairs for summarization or (context, response) pairs for dialogue generation. These pairs are retrieved from the database and serve as demonstration examples to guide a LLM in conducting text generations. The underlying rationale of this paradigm is that better demonstrations typically prompt better generation outcomes.

\smallskip\smallskip\smallskip
\noindent\textbf{Q2. Why applying such partitioning, what intuition behind that, instead of improving the quality of retrieval or introduce more dimensions in the scoring function to account for categories/partitions?}

We recognize that database partitioning plays a crucial role in efficiently managing a database. However, this aspect has been relatively underexplored in the context of RAG, despite the necessity of accessing an external database to obtain essential information for LLM generation. To address this gap, we investigate a multiple partition paradigm for executing RAG. The rationale behind this approach is intuitive: with various attributes associated with the data in a database, queries should ideally be matched with their corresponding attributed data, thereby filtering out noise data.

We discuss our choice of employing partitioning for RAG instead of two alternative approaches: (1) improving the quality of retrieval or (2) introduce more dimensions in the scoring function to account for categories/partitions.

For (1), improving retrieval quality typically emphasizes the effectiveness of AKNN search, often measured using metrics such as recall. However, this focus is not entirely aligned with the primary objective of RAG, which is to generate a good response. In the \texttt{M-RAG} framework, we prioritize the quality of LLM generation as an end-to-end metric explicitly guiding the retrieved information.

For (2), unlike attending to data categories or partitions, we observe that the multiple partition setup offers a cost-effective approach to enhance effectiveness, as confirmed in Figure~\ref{fig:partition}. In this context, no additional computation associated with the LLM is required. Instead, we can keep the LLM frozen, and explore (via Agent-S) or revise (via Agent-R) a relevant memory. This typically leads to improved generation results for the LLM.

\smallskip\smallskip\smallskip
\noindent\textbf{Q3. What is the motivation of the Agent-R and the revision of the retrieved memory?}

\texttt{M-RAG} involves a Retrieval-then-Generation process employing LLMs, typically containing billions of parameters. Here, the LLM remains frozen while the retrieved memories undergo revision before being fed back into the LLM to enhance results. Common revision operations within the retrieved memory, such as re-ranking content~\cite{Blagojevi/Enhancing}, eliminating irrelevant context~\cite{anderson-etal-2022-lingua}, summarizing key information~\cite{DBLP:journals/corr/abs-2310-05029}, and generating candidates for selection~\cite{cheng2023lift}, have been extensively studied in retrieval-augmented literature, as highlighted in the survey paper~\cite{DBLP:journals/corr/abs-2312-10997}. In our work, we conceptualize memory revision as a Markov Decision Process (MDP) and investigate a reinforcement learning solution employing the proposed Agent-R for this operation.

\smallskip\smallskip\smallskip
\noindent\textbf{Q4. \texttt{M-RAG} relies on the partitioning strategy. If the partitions are not well-optimized, it could lead to suboptimal retrieval and generation performance?}

The performance of \texttt{M-RAG} is preserved through several measures. First, we conduct an empirical study, depicted in Figure~\ref{fig:partition}, to investigate a partitioning strategy that outperforms retrieval from the entire database. This serves as a prerequisite for achieving performance improvements. Additionally, building upon this prerequisite, the challenge shifts to identifying suitable partitions and enhancing data quality within them, tasks that are addressed concurrently by two agents. As illustrated in the ablation study presented in Table~\ref{tab:ablation}, performance gains are still attainable even if one agent fails, suggesting that performance improvements can be expected with the \texttt{M-RAG} approach.

\end{document}